# Difficulty Estimation and Simplification of French Text Using LLMs

Henri Jamet, Yash Raj Shrestha, Michalis Vlachos

Dept. of Information Systems, Faculty of Business and Economics (HEC)
University of Lausanne

**Abstract.** We leverage generative large language models for language learning applications, focusing on estimating the difficulty of foreign language texts and simplifying them to lower difficulty levels. We frame both tasks as prediction problems and develop a difficulty classification model using labeled examples, transfer learning, and large language models, demonstrating superior accuracy compared to previous approaches. For simplification, we evaluate the trade-off between simplification quality and meaning preservation, comparing zero-shot and fine-tuned performances of large language models. We show that meaningful text simplifications can be obtained with limited fine-tuning. Our experiments are conducted on French texts, but our methods are language-agnostic and directly applicable to other foreign languages.

## 1 Introduction

Today, there exist many online foreign-language learning tools, including Duolingo, Frantastique, and ReadLang which incorporate gamification elements to increase appeal and retention [25]. However, they do not offer personalization based on learners' interests nor do they provide relevant, contemporary content. Oftentimes, the educational content fails to be matched to the learners knowledge level. Content that is too easy bores learners; too difficult discourages them.

In this work, we use large language models (LLMs) to estimate and potentially reduce the difficulty of foreign content. Such an approach, can be used to identify appropriate contemporary content in the target learning language, thus increasing learners' motivation. Moreover, such an approach could be used in conjunction with a recommendation system to discover content appropriate for the learner's knowledge level. The **contributions** of this work include:

- A machine learning solution to estimate foreign text difficulty, more accurate than traditional *readability* metrics.
- Use of fine-tuned LLMs to simplify the difficulty/level of a foreign French text while preserving its meaning as well as possible. We propose a technique for automatically assessing simplification quality and report performance benchmarks.

## 2 Difficulty Estimation

### 2.1 Related work

From a pedagogical perspective, our solution builds upon already established theory of *extensive reading* [7,15], which constitutes a crucial means of reinforcing one's language skills. However, it is important to find content that is appropriate for the learner's knowledge level of the foreign language. One approach to estimating the difficulty of a foreign language text is to use "readability" formulas[1]. They are regression approaches that calculate the complexity of a text based on various text features, including sentence length and word frequency. Some commonly used readability formulas for foreign language texts include the **Flesch-Kincaid Grade Level**, the **Simple Measure of Gobbledygook** (SMOG), and the **Gunning Fog Index**.

Another approach to tackle difficulty estimation is to use machine learning techniques to predict the difficulty level based on various linguistic features [5,10]. A particularly notable advancement in this field in recent years is the integration of pre-trained word- and sentence- embeddings into text readability architectures [27,12,16,11]. However, our review indicates that prior research has not yet investigated the predictive accuracy of difficulty estimation using LLMs.

### 2.2 Our approach

We model the estimation of difficulty as a classification problem. Let $\mathcal{D}$ be the set of documents, $Y$ the random variable representing the linguistic difficulty class. For the linguistic difficulty, we wish predict the Common European Framework of Reference for Languages, or CEFR difficulty level of a text $\{A1, A2, B1, B2, C1, C2\})$, where A1,A2 signify easy text, B1,B2 intermediate level and C1,C2 advanced level. $f : \mathcal{D} \rightarrow Y$ is a classifier mapping a document $d \in \mathcal{D}$ to a difficulty class $y \in Y$. The classifier $f$ is built using text-label pairs, where the label corresponds to the linguistic difficulty. We use LLMs like **BERT** [8], **GPT** [20], **GPT-3** [2], **LLaMa** [24], and **Palm** [6], which convert text tokens into embeddings capturing the meaning of each token. The models vary in data volume, training methodology, size, and language capabilities, resulting in embeddings of different lengths (e.g., 768 for **BERT**, 1,536 for OpenAI's "ada-002").

**GPT-3+ Models:** Advanced models like **GPT-3** and **GPT-4** [20] have been trained on massive multilingual datasets and post-trained using reinforcement learning from human feedback. They achieve state-of-the-art performance across tasks like summarization, translation, and question-answering [3]. These models excel at generating human-like text and can be fine-tuned for classification tasks. We use **GPT** models for their good performance-versus-cost ratio.

---

[1] Note, that these readability formulas are primarily targeted to estimate the difficulty of a text for native speakers rather than for second language learners [28]. These methods were initially developed for the English language, and progressively have been extended for other languages such as French, Chinese and Italian [19,5].

## 3  Text Simplification

In the context of a foreign language learning app, we now explore the scenario assuming we have discovered content that matches the user's interests (such as, sports or politics), but it is more advanced than their current language proficiency level.

Evaluating text simplification systems is a non-trivial task that requires metrics robust enough to account for both readability and semantic preservation. Traditional n-gram based metrics such as **BLEU** or **ROUGE**, though popular in translation tasks, have limitations when applied to text simplification due to the fundamental requirement of simplification to alter the text structure while maintaining the same meaning [1] [22]. The **SARI** metric, designed specifically for simplification tasks, measures the goodness of words added, deleted, and kept by the system. Current research, such as [22], is exploring new methods like **QUESTEVAL**, which uses semantic questioning of texts to assess simplifications. This approach aims to overcome the limitations of earlier metrics and may result in improved alignment with human evaluations.

While text summarization and text simplification may appear similar as both involve altering the original text, their objectives are distinct. Summarization aims to condense a text by trimming down its length and retaining only the main points. Simplification seeks to lower the linguistic complexity, making content more accessible [22]. Simplification may or may not alter the length of text. Unlike summarization, simplification is deals with adjusting the text to a particular knowledge or comprehension level.

The hybrid task of combining both text summarization and simplification has been investigated in the context of generating summaries for lay audiences. This task known as "lay summarization" aims to render complex scientific content accessible to a general audience, requiring the extraction of key points and simultaneously a reduction in linguistic complexity [26,4]. Transformer models based on **BERT** [8] and **PEGASUS** [29] have been used in this context.

Recent advancements in LLMs, have greatly contributed to the field of text simplification [21]. However, the efficacy of these models varies, and not all are equally suited for the task, calling for evaluation and fine-tuning on specialized datasets [23]. A recent analysis by [9] highlights that when evaluated on public datasets, contemporary LLMs like **GPT-3** can match or even outperform systems explicitly designed for text simplification.

### 3.1  Our approach

We model simplification in a similar way as for content difficulty estimation. However, instead of predicting the level of difficulty, we predict the simplified sentence token by token. We provide labelled examples of original and simplified sentences and train a machine learning model. Because we only fine-tune LLMs, we need to provide very few pairs of examples, since the LLMs have already encapsulated in them large amounts of textual knowledge. In our experiments, we only post-train the LLMs using 125 pair sentences, and show large improvements compared to the zero-shot LLM counterpart. Because our goal is to assist learners in improving their language skills by reading content close to their knowledge/experience level, simplification of a given text is approached as a sentence-by-sentence simplification, and not as a complete synthesis of new text.

A key challenge in text simplification involves evaluating and balancing the trade-off between preserving the meaning and semantics of the original text and achieving effective simplification. The more we simplify, the more we may lose the original meaning, depending on the complexity of the original text. To evaluate these conflicting aspects, we introduce two metrics: simplification accuracy, and semantic similarity. Therefore, given a set of original sentences $\mathcal{O}$, their simplified versions $\mathcal{T}$ by a model $m$ that we seek to evaluate $\mathcal{O} \xrightarrow{m} \mathcal{T}$, and $\epsilon(\cdot)$ is the embedding representation of a text, we define: **A. Simplification Accuracy** $A \in \{0, 1\}$ is a binary output indicating if a text $t \in \mathcal{T}$ is exactly one CEFR level of difficulty lower than their associated original text $o \in \mathcal{O}$. As an example, if a text of level C2 is simplified to C1 with a given model then $A = 1$, but if the resulting simplification leads to B2 level text, then $A = 0$. For a set of texts, we average $A$. **B. Semantic Similarity** $S \in [0\ldots1]$ is a real number between zero and one, and captures the preservation of semantic meaning of simplified text with respect to original text, which can be measured as the cosine similarity between the embedding representation $\epsilon(\cdot)$ of the original text and the embedding representation of the simplified text. The choice of model used to calculate embeddings is described below. For a set of texts, we average $S$. Note, that this measure of similarity allows texts to have completely different words, but still have high similarity if their vectors are close in the (semantic) embedding space. We integrate both aspects, simplification and semantic similarity, into a single weighted-score by drawing inspiration from the *F1-score*:

$$\text{w-Score} = 2 \times \frac{w_1 \times A \times w_2 \times S}{w_1 \times A + w_2 \times S}$$

where $w_2 = (1 - w_1)$. The coefficients $w_1$ and $w_2$ are chosen in such a way to balance the importance of the two aspects (equal in our experiments). This metric allows us to compare the performance of each of the models evaluated. Note that the $A$ and $S$ components would benefit from being kept separate to easily distinguish the strengths and weaknesses of each model.

There is one fine aspect still to be discussed. While $S$ can be easily computed, evaluating the simplification accuracy $A$ is more challenging. Even though we have the difficulty of the original text, when a text is simplified by a model $m$, then we do not have at our disposal the difficulty level of the simplified text (A1-C2). In our experiments, we address this by using the fine-tuned **CamemBERT** model as the *proxy* evaluator of the simplification accuracy. We compute the difficulty of the original text and the simplified text, and if the difficulty is reduced by one level, we consider the transformation as valid. For example, if the original label was C2, and **CamemBERT** classified it as C1, but then it classified the simplified text as B2, then we assign $A = 1$ since the difficulty reduction was one level. Notice, that this approach should cater for constant biases or potential errors that the model might introduce.

## 4  Experiments

The goal of the following experiments is to demonstrate that (a) fine-tuned LLMs can significantly improve the difficulty estimation offered by traditional readability metrics. (b) simplification methodology driven by fine-tuned LLM models outperforms zero-shot approaches. The code for the experiments can be retrieved here .

### 4.1  Difficulty estimation

We evaluated the difficulty estimation on three labeled **datasets**: **1)** Littérature de jeunesse libre (`LjL`) which we obtained from [11]. Each content item here contains several sentences and a label (labels: level1, level2, level3, level4). **2)** A collection of sentences collected from the Internet (`sentencesInternet`). Each of these sentences was then annotated by at least three annotators (students recruited in our university) in difficulty levels. Only sentences in which all participants agreed on the difficulty annotation were retained (labels: A1,A2,B1,B2,C1,C2). Here, the labels correspond to the levels designed by the Common European Framework of Reference for Languages (CEFR). **3)** A collection of sentences from literature books (`sentencesBooks`). Each book was annotated with a difficulty level by a Professor of French. All sentences in that book were then given that label. This process involved an OCR pipeline which could lead to faulty detection of characters, so only the sentences without any errors were retained. (labels: A1,A2,B1,B2,C1,C2). The characteristics of these datasets are provided in Table 1. To train and evaluate our model, we used an 80/20 train-test split, and the results that we present are for examples which the model saw for the first time.

Table 1: Characteristics of datasets for difficulty estimation experiments

| Dataset | Sentences | Words | Chars | Labels |
|---|---|---|---|---|
| `LjL` [11] | 2,060 | 334,026 | 1,532,442 | level1–4 |
| `sentencesInternet` | 4,800 | 85,941 | 421,045 | A1–C2 |
| `sentencesBooks` | 2,400 | 56,557 | 281,463 | A1–C2 |

**Evaluation.** As a simple benchmarking, we compare the accuracy of our difficulty estimation approach to traditional readability-based metrics, such as the **GFI** (Gunning Fog Index), **FKGL** (Flesch Kincaid grade level), **ARI** (Automated Readability Index) in Table 2. Initially, all of these metrics have been developed for English content, but language specific models, such as for French text, as used in our application, have also been developed [10].

These techniques are inherently regression techniques and output a floating point value of the text difficulty. As a result, we cannot make a direct comparison, because our difficulty estimator predicts discrete labels. To address this challenge, we trained a logistic regression model $\mathbb{R} \to \mathcal{L}$ with $\mathcal{L}$ being the space of our labels equal to $\{\text{level1}, \text{level2}, \text{level3}, \text{level4}\}$ for the LjL dataset and $\{A1, A2, B1, B2, C1, C2\}$ for the `sentencesBooks` and `sentencesInternet` datasets. In this way, we transformed our regression into a classification, which we can evaluate with the usual metrics and compare with the other classification methods. For LLM-based difficulty estimation, we train the following classifiers using the examples from the training set and test how they behave on the unseen test set: **1. GPT**-based models: We use the **GPT-3.50-turbo-1106** and **Davinci-002**. Both these models have been fine-tuned for the task at hand using the examples with the labelled difficulties of the text. **2. CamemBERT**: This is essentially the well-known **BERT** model trained on French data [18]. **CamemBERT** is based on the **RoBERTa** architecture [17]. While **BERT** was initially trained on a diverse range of texts in multiple languages, **CamemBERT** is specifically fine-tuned for the French language. This fine-tuning process involves training the model on a massive amount of French text data, which enables it to capture the nuances, idioms, and syntactic structures unique to the French language. Because **CamemBERT** is tailored to French, it excels in various language-related tasks such as text classification, sentiment analysis, and named entity recognition within the context of French text. **3. Mistral-7B**: This is an open-source LLM [13] with a modest size that we fine-tuned in the same manner as GPT-based models.

To take advantage of the inherent knowledge in LLMs, we tested different contexts and selected the one which offered the best performance. We then evaluated which models could potentially benefit with and without this context.[2] Table 2 shows the F1-score across the three datasets. The column *context* indicates whether the model has been trained with or without a task-related context. The various models were sorted by average ranking of their F1-score on each dataset. Our results suggest that the **GPT-3.5** is the best performing model. The second place is shared by the **Mistral-7B** and the **CamemBERT**, but we draw the attention of the reader to the much smaller size of these models: 7 Billion parameters for **Mistral-7B**, 110 Million for **CamemBERT**, compared to 175 B parameters of **GPT-3.5**.

### 4.2   Text simplification

For the training and the evaluation of the LLM models, we have constructed two different datasets: **1) Training-set.** To fine-tune our models for the task of simplification, we need a dataset of French sentences with their simplifications at an associated lower CEFR level. We used **GPT4** to generate 125 sentences (*25 from each level `A2, B1, B2, C1, C2`*) and their simplified versions. This dataset was further reviewed by a native French speaker. **2) Test-set.** We take, per difficulty level `A2, B1, B2, C1, C2` (*Level A1 cannot be simplified*), 100 random sentences from the `sentencesBooks` and `sentencesInternet` dataset. The test-set consists of $5 \times 100 \times 2 = 1000$ sentences.

---

[2] The context used consists of a French text whose translation is given by: *You are a language assessor using the Common European Framework of Reference for Languages (CEFR). Your task is to assign a language proficiency score to this text, using the CEFR levels from A1 (beginner) to C2 (advanced/native). Evaluate this text and assign it the corresponding CEFR score.*

Table 2: Difficulty estimation metrics for all datasets

| model | context | LjL Sentences | Internet Sentences | Books |
|---|---|---|---|---|
| GPT-3.5-turbo-1106 | ✓ | 0.72 | **0.90** | 0.50 |
|  | - | **0.73** | 0.87 | 0.49 |
| CamemBERT | - | 0.62 | 0.82 | **0.52** |
| Mistral-7B | ✓ | 0.64 | 0.75 | 0.51 |
| Davinci-002 | - | 0.59 | 0.82 | 0.47 |
|  | ✓ | 0.61 | 0.81 | 0.47 |
| Mistral-7B | - | 0.47 | 0.63 | 0.35 |
| FKGL | - | 0.42 | 0.34 | 0.35 |
| GFI | - | 0.45 | 0.32 | 0.34 |
| ARI | - | 0.40 | 0.34 | 0.34 |

**Model Evaluation.** We examine the performance of **GPT-4**, **Davinci** and **GPT-3.5-turbo-1106** from OpenAI, and the **Mistral-7B** model. We provide the results of our evaluation in Table 3. Among the different models evaluated, GPT-4 Zero-shot has the highest w-Score. The 0.5 in the simplification accuracy of GPT-4 shows that in 50% of the cases, the text was simplified to one-level lower of difficulty. At the same time the meaning is highly preserved with the cosine similarity between the original and simplified embeddings of the text being 0.89 on average.

Table 3: Performance of LLMs for the text simplification task.

| Model | Simplification Accuracy | Semantic Similarity | w-Score |
|---|---|---|---|
| **GPT-4 Zero-shot** | *0.50* | *0.89* | *0.64* |
| **Mistral-7B Fine-Tuned** | 0.35 | 0.91 | 0.51 |
| **GPT 3.5 Fine-Tuned** | 0.34 | 0.91 | 0.50 |
| **GPT 3.5 Zero-Shot** | 0.31 | **0.93** | 0.47 |
| **Mistral-7B Zero-Shot** | 0.28 | 0.93 | 0.43 |
| **Davinci Fine-Tuned** | 0.24 | 0.83 | 0.38 |

**Iterative Simplification.** Finally, we illustrate how a simplification model, the Mistral in this case, behaves for the task of an iterative simplification of French sentences. We randomly selected 100 level `C2` sentences from the `training-set` correctly classified by **CamemBERT** as being of level `C2`. We then iteratively applied the simplification steps to the sentence using the fine-tuned version of **Mistral-7B** before evaluating the difficulty of the resulting sentence and the cosine similarity with the original sentence (from iteration 0). For a performance model, that always reduces the difficulty one-difficulty level at a time, we could only do this operation 5 times (`C2` ⟶ `A1`). Since, we don't have a perfect model, we perform this 8 times to observe the trend. We see that the model successfully reduces both the difficulty and the semantic similarity is also reduced as a consequence of the simplification, as shown in Figure 1.

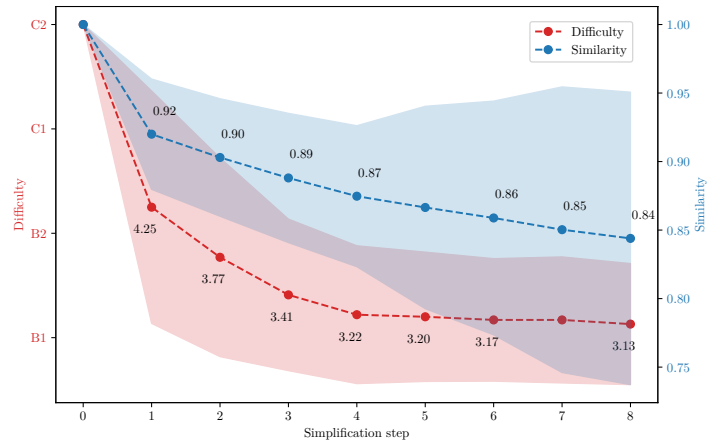

Fig. 1: Iterative simplification of sentence difficulty of CEFR level C2 with Mistral-7B model where CamemBERT was used to estimate difficulty and cosine similarity to evaluate text similarity. We report average value over 100 experiments.

## 5   Conclusion

This study demonstrates the potential of LLMs to enhance the estimation of foreign text difficulty and simplification. These advancements open new perspectives for personalizing language learning. By integrating these models into educational platforms, it becomes possible to adapt content to each learner's interests and level, making the experience more engaging and effective. Moreover, these models could help bridge gaps in existing pedagogical resources by generating simplified content on-demand. Future work should explore the possibility of working with entire paragraphs rather than isolated sentences for difficulty estimation and simplification. It would also be interesting to include state-of-the-art models like **GPT-4**, **Claude 3 Opus**, **Gemini 1.0 Ultra**, and larger open-source models such as **Mistral 8x22b** in the comparisons [14].